\pdfoutput=1

\documentclass[11pt]{article}

\usepackage[final]{acl}

\usepackage{times}
\usepackage{latexsym}

\usepackage[T1]{fontenc}

\usepackage[utf8]{inputenc}

\usepackage{microtype}

\usepackage{inconsolata}

\usepackage{graphicx}

\usepackage{amsmath}
\usepackage{amssymb}
\usepackage{mathtools}
\usepackage{amsthm}
\usepackage{enumitem}
\usepackage{multirow}
\usepackage{booktabs}
\usepackage{algorithm}
\usepackage{algorithmic}
\usepackage{bbding}
\usepackage{placeins} 

\title{AVG-LLaVA: An Efficient Large Multimodal Model with  \\ Adaptive Visual Granularity}

\author{
 \textbf{Zhibin Lan\textsuperscript{1,3}}\thanks{~~Work was done when Zhibin Lan was interning at Pattern Recognition Center, WeChat AI, Tencent Inc, China.},\quad
 \textbf{Liqiang Niu\textsuperscript{2}},\quad
 \textbf{Fandong Meng\textsuperscript{2}},\quad
 \textbf{Wenbo Li\textsuperscript{1,3}},\quad
 \textbf{Jie Zhou\textsuperscript{2}},\quad
 \textbf{Jinsong Su\textsuperscript{1,3,4}}\thanks{~~Corresponding author.}
\\
 \textsuperscript{1}School of Informatics, Xiamen University, China,\\
 \textsuperscript{2}Pattern Recognition Center, WeChat AI, Tencent Inc, China, \\
 \textsuperscript{3}Key Laboratory of Digital Protection and Intelligent Processing of Intangible Cultural Heritage \\ of Fujian and Taiwan (Xiamen University), Ministry of Culture and Tourism, China, \\
  \textsuperscript{4}Shanghai Artificial Intelligence Laboratory, China \\
 \small{
   {lanzhibin@stu.xmu.edu.cn,\quad jssu@xmu.edu.cn}
 }\\
}

\begin{document}
\maketitle
\begin{abstract}
Recently, large multimodal models (LMMs) have achieved significant advancements. When dealing with high-resolution images, dominant LMMs typically divide them into multiple local images and a global image, leading to a large number of visual tokens. In this work, we introduce AVG-LLaVA, an LMM that can adaptively select the appropriate visual granularity based on the input image and instruction. Specifically, we first apply the multiple pooling layers to obtain visual tokens at different granularities. Then we propose a visual granularity router, which includes a Transformer layer, an MLP layer, and a voter layer, used to select the appropriate visual granularity based on the image and instruction. Furthermore, we put forward RGLF, a novel training paradigm that aims at aligning the granularity predicted by the router with the preferences of the LMM, without the need for additional manually annotated data. Extensive experiments and analysis show that AVG-LLaVA achieves superior performance across 11 benchmarks, as well as significantly reduces the number of visual tokens and speeds up inference (e.g., an 85.3\% reduction in visual tokens and a 2.53$\times$ increase in inference speed on the AI2D benchmark). Our code and model can be found at \url{https://github.com/DeepLearnXMU/AVG-LLaVA}.
\end{abstract}

\section{Introduction}
Recently, the field of artificial intelligence (AI) has witnessed a significant advancement in large multimodal models (LMMs) \citep{gpt-4v,DBLP:journals/corr/abs-2304-10592,DBLP:conf/nips/InstructBLIP,DBLP:conf/nips/LiuLWL23a,Liu_2024_CVPR}, marking a further step toward artificial general intelligence (AGI). 
Most existing LMMs follow the structure of LLaVA \citep{DBLP:conf/nips/LiuLWL23a,Liu_2024_CVPR}, which includes a vision encoder to embed input images into visual tokens and a connector to map them into the word embedding space. Subsequently, these visual tokens are fed into a large language model (LLM) \citep{DBLP:journals/corr/abs-2302-13971, DBLP:journals/corr/abs-2303-08774, vicuna2023} for multimodal understanding and reasoning \citep{DBLP:conf/icml/BLIP2,liu2023retrieval, DBLP:journals/tmlr/0001Z00KS24,lin2025investigating}, alongside the word embeddings.

\begin{figure}[t]
    \centering
    \includegraphics[width=1\linewidth]{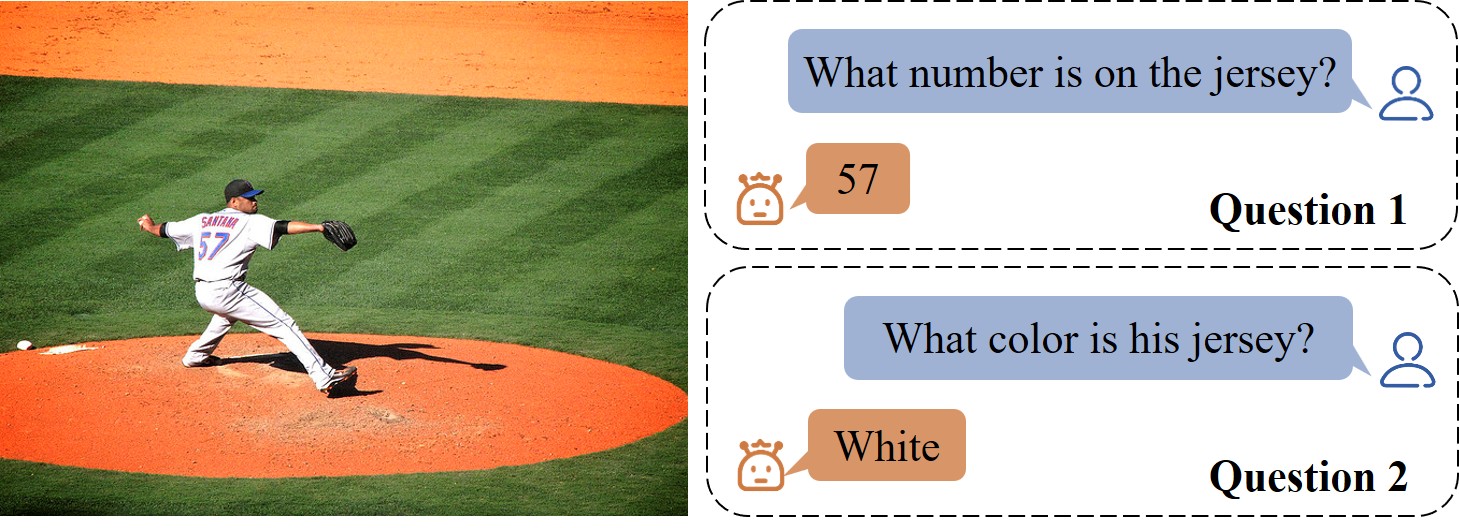}
    \caption{An example of VQA from MSCOCO \citep{DBLP:conf/eccv/coco}. Notably, responding to Question 1 necessitates fine-grained visual information, whereas responding to Question 2 requires only coarse-grained visual information.}
      \label{fig:example}
    \vspace{-4mm}
\end{figure}

Due to the limitations imposed by the fixed aspect ratio (e.g., 1:1) and low resolution (e.g., 336×336) used by visual encoders (e.g., CLIP-ViT \citep{DBLP:conf/icml/CLIP}), early LMMs face challenges in processing high-resolution images with different aspect ratios. To deal with this limitation, dominant models, such as LLaVA-NeXT \citep{liu2024llavanext}, dynamically divide each input high-resolution image into multiple local images. These local images are encoded separately, and their tokens are then concatenated with the tokens of the original global image. This approach will lead to longer visual token sequences, such as 2880 visual tokens for a 672×672 image. However, in practice, such fine-grained visual information is not always necessary, and in some cases, coarse-grained visual information can even be more beneficial for model predictions. For instance, as shown in Figure \ref{fig:example}, when the model is asked to recognize the number on the jersey, it requires relatively fine-grained visual information. In contrast, determining the color of the jersey only necessitates coarse-grained visual information.

In this paper, we propose \textit{Adaptive Visual Granularity LLaVA (AVG-LLaVA)}, an LMM that can adaptively select the appropriate visual granularity based on the input image and instruction. The basic intuition behind our model is that humans only scrutinize images carefully when answering difficult questions; otherwise, a brief glance is sufficient.

\begin{figure*}[t]
    \centering
    \includegraphics[width=1\linewidth]{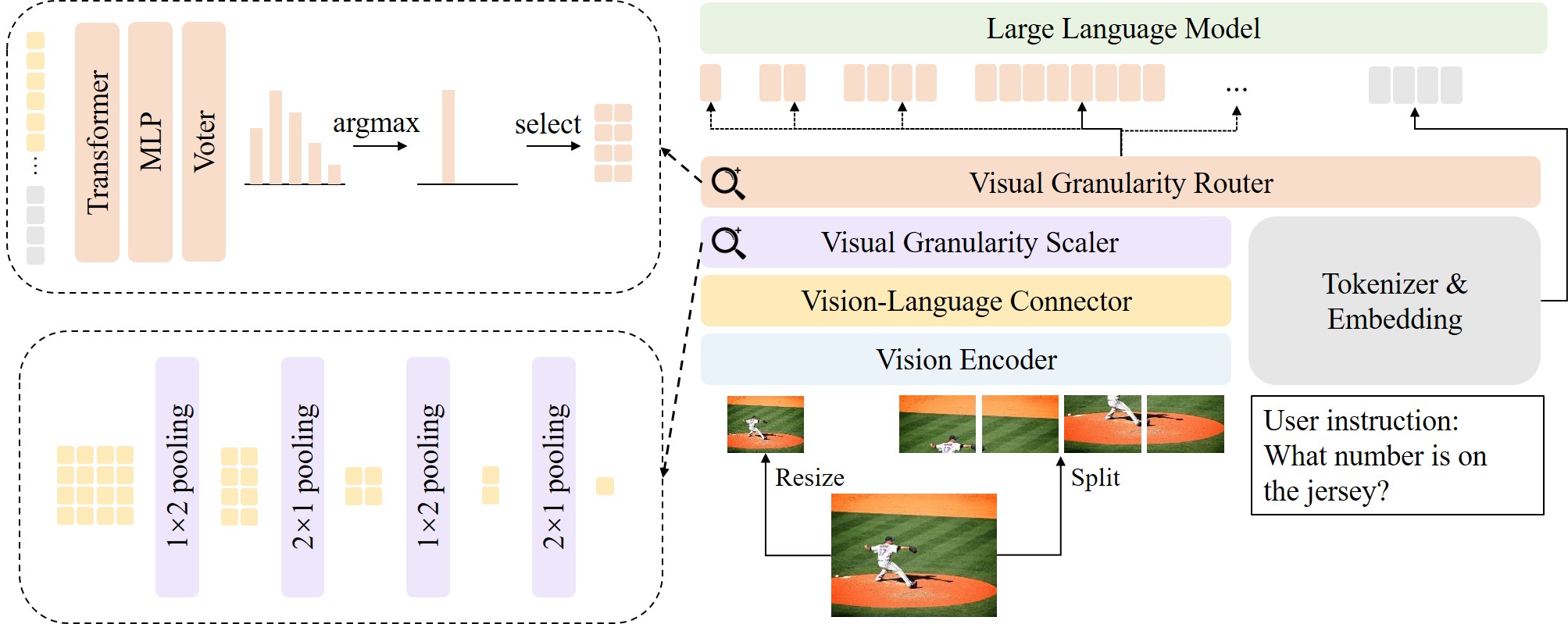}
    \caption{The architecture of AVG-LLaVA. AVG-LLaVA additionally introduces two modules based on LLaVA-NeXT: (1) Visual granularity scaler. This module consists of multiple pooling layers that progressively increase the granularity of visual features, thereby reducing the number of visual tokens; (2) Visual granularity router. This module includes a Transformer layer, an MLP layer, and a voter layer, which are used to select the appropriate granularity of visual features based on the input multi-granularity visual tokens and instruction tokens.}
    \label{fig:model}
\end{figure*}

As displayed in Figure \ref{fig:model}, AVG-LLaVA extends LLaVA-NeXT with a \textit{visual granularity scaler} and a \textit{visual granularity router}. The visual granularity scaler performs multiple rounds of pooling on visual tokens, each time halving the number of visual tokens, thus obtaining a series of visual features with different granularities. The visual granularity router adaptively selects the appropriate visual granularity features based on the input multi-granularity visual features and text features. By doing so, for images and instructions that do not require fine-grained details, the number of visual tokens can be reduced, which not only speeds up inference but also may improves performance. 
This performance enhancement likely stems from the reduction of redundant information, as selecting appropriate visual granularity makes it easier for the model to answer questions based on images effectively.


Besides, we observe that it is challenging to train the visual granularity router directly through visual instruction tuning \citep{DBLP:conf/nips/LiuLWL23a}. This may be because the router cannot learn the distinctions between different visual granularities from visual instruction tuning, making it difficult to learn how to select the most appropriate visual granularity based on the image and instruction. 
To deal with this issue, we propose a novel training paradigm called \textit{Ranking} \textit{Granularity} based on \textit{LMM} \textit{Feedback} (\textit{RGLF}). This paradigm aligns router probabilities of multiple granularities with LMM preferences by a ranking loss \citep{DBLP:conf/emnlp/HopkinsM11,DBLP:conf/acl/BRIO}, effectively aiding the router in distinguishing between different visual granularities and selecting the appropriate one.

We further evaluate AVG-LLaVA on 11 benchmarks including tasks from various types (e.g., general VQA and text-oriented VQA, etc.). Extensive experimental results show that AVG-LLaVA can effectively reduce the number of visual tokens and improve inference speed (e.g., an 85.3\% reduction in visual tokens and a 2.53$\times$ increase in inference speed on the AI2D \citep{DBLP:conf/eccv/ai2d} benchmark) while achieving better performance under the same base LLM.

\section{Related Work}
\paragraph{High-Resolution LMMs.} 
Large language models (LLMs) such as GPT-4 \citep{DBLP:journals/corr/abs-2303-08774}, LLaMA \citep{DBLP:journals/corr/abs-2302-13971}, and Gemini \citep{team2023gemini} have achieved significant success in language understanding and generation, driving the development of LMMs that integrate vision encoders with LLMs and leverage visual instruction data for fine-tuning. However, early LMMs \citep{DBLP:conf/icml/BLIP2,DBLP:journals/corr/LLaMA-Adapter,DBLP:conf/nips/LiuLWL23a} rely on fixed-resolution (e.g., 336×336) CLIP-ViT to process images, which limits their ability to capture high-resolution image details.

To perceive images with higher resolutions, Qwen-VL \citep{qwen-vl} increases the input resolution of the visual encoder to 448×448 and introduces an additional training stage. Along this line, both Vary \citep{DBLP:journals/corr/vary} and Mini-Gemini \citep{DBLP:journals/corr/Mini-Gemini} include two vision encoders: one is an additional introduced high-resolution vision encoder, and the other is the original low-resolution vision encoder. Unlike the methods mentioned above, SPHINX \citep{DBLP:journals/corr/SPHINX} and Monkey \citep{Monkey} enlarge the input image to a high resolution, and then divide it into a fixed number of local images, which are individually encoded using an image encoder to obtain local image tokens. Subsequently, the original global image tokens are concatenated with all local image tokens to feed into the LLM.
Furthermore, LLaVA-NeXT \citep{liu2024llavanext} enumerates various resolutions and adaptively selects the one that most closely matches the input image resolution. Although these methods can achieve better performance, they significantly increase the number of visual tokens, as the computational complexity scales quadratically with the number of input tokens, resulting in higher inference costs.

\paragraph{Vision Token Reduction for LMMs.}
Recently, several methods are proposed to reduce the visual tokens for LMMs, including visual token compression and pruning. For example, LLaVA-UHD \citep{LLaVA-UHD} adopts a QFormer-like \citep{DBLP:conf/nips/InstructBLIP} structure to compress visual tokens, while MG-LLaVA \citep{DBLP:journals/corr/abs-2406-17770} employs a convolution layer to compress high-resolution visual features. In contrast, CrossGET \citep{DBLP:journals/corr/CrossGET} introduces a cross-modal token for leveraging cross-modal information to make decisions on token selection and merging. LLaVA-PruMerge \citep{DBLP:journals/corr/LLaVA-PruMerge} employs the similarity between the class token and other tokens as a key criterion for pruning and merging vision tokens. 

Furthermore, FastV \citep{DBLP:journals/corr/FastV} finds that most image tokens receive inefficient attention after the second decoder layer, and thus prunes half of the image tokens. 
Similarly, VTW \citep{DBLP:journals/corr/VTM} adopts a more aggressive strategy to prune all visual tokens at a certain layer. 
Unfortunately, despite the above methods effectively reducing the number of visual tokens, they often lead to a certain degree of decline in model performance. More recently, LLaVA-\textit{M}$^3$ \citep{DBLP:journals/corr/m3} obtains multi-granularity visual features by merging visual tokens through pooling, enabling manual control of the tradeoff between inference cost and performance. 

Significantly different from aforementioned methods, our model is a dynamic neural network \citep{DBLP:journals/pami/HanHSYWW22} that can adaptively select the appropriate visual granularity based on the input image and instruction, improving model performance while reducing the number of visual tokens.

\section{Our Model}
\subsection{Model Architecture}

As shown in Figure \ref{fig:model}, in addition to the visual encoder, visual-language connector, and LLM, AVG-LLaVA introduces two additional modules on top of LLaVA-NeXT: the visual granularity scaler and the visual granularity router. The key components will be elaborated in the following.

\paragraph{High-Resolution Image Encoding.} Given an input image $\mathbf{I}$ $\in$ $\mathbb{R}^{H \times W \times 3}$, we follow common practice \citep{liu2024llavanext} to divide it into multiple smaller local images $\mathbf{I}_{local} \in \mathbb{R}^{H_v \times W_v \times 3}$. Here, $H_v$ and $W_v$ are the resolution that the vision encoder is originally trained for. Then, these local images are individually encoded into a $H_p \times W_p$ grid of visual tokens $\mathbf{X}_{local} \in \mathbb{R}^{H_p \times W_p \times C}$ by the image encoder, where $C$ is the dimension of the visual encoder. To preserve the global context information of the input image, we resize the original image to $H_v \times W_v$) and encode it as global visual tokens. 
Finally, we map both global and local visual tokens to the word embedding space through an MLP-based vision-language connector.

\label{sec:vgs}
\paragraph{Visual Granularity Scaler.} This module follows the design of spatial pyramid pooling \citep{DBLP:journals/pami/HeZR015,DBLP:journals/corr/m3}, sequentially stacks 1$\times$2 and 2$\times$1 average pooling layers, thereby obtaining visual features at multiple granularities and preserving the spatial information. In this work, we consider CLIP-ViT-L-336 \citep{DBLP:conf/icml/CLIP} as the visual encoder, and thus each image is encoded into 24$\times$24 grid of visual tokens. Then, these visual tokens are fed into the visual granularity scaler, obtaining visual tokens with a grid of 24$\times$12, 12$\times$12, 12$\times$6 and 6$\times$6, respectively. In this way, we can obtain visual tokens of different granularities in a fine-to-coarse manner without training.

\paragraph{Visual Granularity Router.} Different visual granularity features can be considered as different experts, so the Mixture of Experts (MoE) \citep{DBLP:conf/iclr/ShazeerMMDLHD17, DBLP:conf/iclr/KomatsuzakiPLRM23,DBLP:journals/corr/MOE-LLaVA,zhang2025advancing} structure is particularly well-suited for selecting the appropriate visual granularity. Unlike the previous MoE studies that use linear layers as routers, we propose a multi-layer structure as illustrated in Figure \ref{fig:model} to select the appropriate visual granularity based on the input image and the instruction.  

Specifically, when dealing with an image, we first flatten and concatenate its visual tokens of all granularities to form multi-granularity visual tokens $\mathbf{\overline{X}}_v = [\mathbf{X}_{v}^1;\mathbf{X}_{v}^2;...;\mathbf{X}_{v}^N]$, where $\mathbf{X}_v^i$ represents the visual tokens of the $i$-th granularity, and $N$ is the number of visual granularities\footnote{To simplify the explanation, we use a single image as an example. In practice, we include a global image and multiple local images, and each image will go through the following steps. The final result will be obtained by averaging the results of all the images.}. Then, these visual tokens are concatenated with the filtered instruction tokens $\mathbf{\overline{X}}_{instruct}$ to serve as the input for the visual granularity router. Here, $\mathbf{\overline{X}}_{instruct}$ is obtained by calculating the cosine similarity between the original instruction tokens $\mathbf{X}_{instruct}$ and the visual tokens with original granularity $\mathbf{X}_{v}$, retaining the top-\textit{k} most relevant ones.
Afterwards, we apply a single Transformer \citep{DBLP:conf/nips/VaswaniSPUJGKP17} layer to facilitate the fusion of visual tokens at different granularities with instruction tokens. Subsequently, an MLP is applied to each token for predicting the appropriate visual granularity, resulting in the logits $\mathbf{Z}_{out}$ $\in$ $\mathbb{R}^{L \times N}$, where $L$ is the number of both visual and instruction tokens. To vote for the most appropriate visual granularity, we use a learnable weight matrix (Voter) $\mathbf{W}$ $\in$ $\mathbb{R}^{1 \times L}$ to aggregate the logits predicted by all tokens, yielding the final logits $\mathbf{Z}_{final} \in \mathbb{R}^{1 \times N}$.
Finally, we use softmax to calculate the probability distribution of visual granularities, where the visual tokens corresponding to the granularity with the highest probability are fed into the LLM.

\subsection{Multi-stage Training}
We provide a detailed description of the training procedures for AVG-LLaVA, which consists of two stages. The first stage endows the model with the ability to perceive and process multi-granularity visual information, while the second stage enables the model to select the appropriate granularity based on the image and instructions.



\paragraph{Stage 1: Multi-Granularity Visual Instruction Tuning.} In this stage, we use the high-quality visual instruction data to train the visual encoder, vision-language connector, and LLM, enabling them to perceive and process visual features of $N$ different granularities. Specifically, we perform next-token prediction using visual features of different granularities and apply the cross-entropy loss only to the answering part, formulated as
\setlength\abovedisplayskip{3pt}
\setlength\belowdisplayskip{3pt}
\begin{equation}
    \mathcal{L}_{1} = - \frac{1}{N} \sum_{i=1}^{N} \sum_{t=1}^{T} \mathrm{log}P(x_t | \mathbf{X}_v^i,\mathbf{X}_{instruct},\mathbf{X}_{a,<t}),
\end{equation}
where $\mathbf{X}_{a}$ are the answer tokens before the current prediction token $x_t$, and $T$ is the length of answer tokens.

\paragraph{Stage 2: Ranking Granularity Based on LMM Feedback.} Then, we introduce the visual granularity router into the model training, where all other modules are frozen, and only the router is trained. This stage allows the model to select the appropriate visual granularity based on the input image and instruction. Intuitively, a straightforward approach to training the router is visual instruction fine-tuning. However, we find that the router trained with this method performs poorly. This could be due to the difficulty of visual instruction fine-tuning in effectively enabling the router to learn the differences between different visual granularities.

\begin{figure}[t]
    \centering
    \includegraphics[width=1\linewidth]{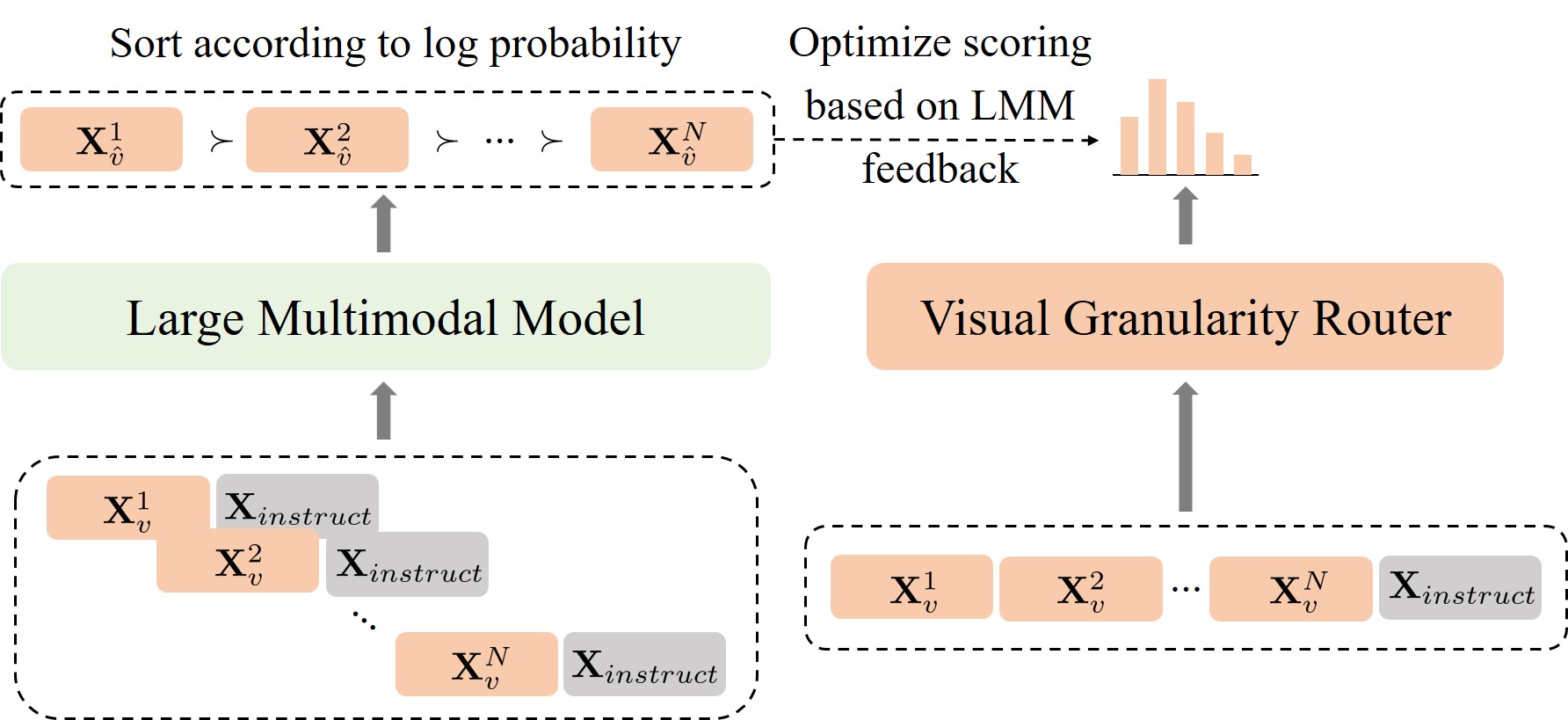}
    \caption{The overview of RGLF. Visual tokens of each granularity are concatenated with instruction tokens and then processed by the LMM to estimate the corresponding rewards. Visual granularity router optimizes the granularity selection based on the feedback from the LMM.}
    \label{fig:RGLF}
\end{figure}

To address the above issue, we propose RGLF, as illustrated in Figure \ref{fig:RGLF}, where the router is trained with a ranking loss, utilizing the feedback from the LMM fine-tuned with multi-granularity visual instructions as the ranking criterion. Concretely, for the given image and instructions, we let the LMM predict answers using visual tokens of different granularity and calculate their respective log probabilities. Then, based on these log probabilities, we sort $\mathbf{X}_{v}^1;\mathbf{X}_{v}^2;...;\mathbf{X}_{v}^N$ in a descending order to obtain $\mathbf{X}_{\hat{v}}^1;\mathbf{X}_{\hat{v}}^2;...;\mathbf{X}_{\hat{v}}^N$. Given the visual tokens $\mathbf{X}_{\hat{v}}^i$ of the $i$-th granularity, we directly consider those tokens ($\mathbf{X}_{\hat{v}}^1$; $\mathbf{X}_{\hat{v}}^2$; ...; $\mathbf{X}_{\hat{v}}^{i-1}$) ranked above it as positive examples and the remaining tokens ($\mathbf{X}_{\hat{v}}^{i+1}$; $\mathbf{X}_{\hat{v}}^{i+2}$; ...; $\mathbf{X}_{\hat{v}}^{N}$) as negative ones. Afterwards, we use the router to give scores (log probability) $s_i$ for each $\mathbf{X}_{\hat{v}}^i$:
\begin{equation}
    s_i = \mathrm{log}P(g_i | \mathbf{\overline{X}}_v, \mathbf{\overline{X}}_{instruct}),
\end{equation}
where $g_i$ denotes the $i$-th granularity predicted by the router based on multi-granularity visual tokens $\mathbf{\overline{X}}_v$ and filtered instruction tokens $\mathbf{\overline{X}}_{instruct}$. Since we expect the router to assign higher probabilities to more appropriate visual granularities, the ranking loss is defined as follows:
\begin{equation}
    \mathcal{L}_{rank} = \sum_{i=1} \sum_{j>i} \mathrm{max}(0, s_j - s_i + \lambda_{ij}),
\end{equation}
where $\lambda_{ij}$ is the log probability difference between the answers predicted by the LLM using visual tokens of the $i$-th and $j$-th granularities:
\begin{equation}
\begin{split}
    \lambda_{ij} = \frac{j-i}{|T|} & \sum_{t=1}^{T} (\mathrm{log}P(x_t | \mathbf{X}_{\hat{v}}^i,\mathbf{X}_{instruct},\mathbf{X}_{a,<t})\\&-\mathrm{log}P(x_t | \mathbf{X}_{\hat{v}}^j,\mathbf{X}_{instruct},\mathbf{X}_{a,<t})).
\end{split}
\end{equation}
When the preference of $\mathbf{X}_{\hat{v}}^j$ is only slightly worse than $\mathbf{X}_{\hat{v}}^i$, the margin will be small. Conversely, when $\mathbf{X}_{\hat{v}}^j$ is significantly worse than $\mathbf{X}_{\hat{v}}^i$, the margin will correspondingly increase. In this way, we can dynamically adjust the margin to obtain adaptively penalty degrees between different pairs.

In addition to aligning with the LMM preference ranking, it is also desirable for the router to select the optimal visual granularity.
Therefore, we add a cross-entropy loss to let the router learn the prediction of granularity with the highest log probability from the LMM, defined as follows:
\begin{equation}
    k = \arg \max \limits_{i} \sum_{t=1}^{T} \mathrm{log}P(x_t | \mathbf{X}_{v}^i,\mathbf{X}_{instruct},\mathbf{X}_{a,<t}),
\end{equation}
\begin{equation}
    \mathcal{L}_{ce} = -\mathrm{log}P(g_k|\mathbf{\overline{X}}_v, \mathbf{\overline{X}}_{instruct}).
\end{equation}
Finally, the total loss is defined as the weighted sum of two losses:
\begin{equation}
    \mathcal{L}_{2} = \mathcal{L}_{rank} + \alpha \mathcal{L}_{ce},
\end{equation}
where $\alpha$ is the hyperparameter used to maintain the balance between the ranking loss $\mathcal{L}_{rank}$ and cross-entropy loss $\mathcal{L}_{ce}$.

\section{Experiments}
\subsection{Settings}

\begin{table*}[t!]
\renewcommand\arraystretch{1.2}
\small
\fontsize{8pt}{8pt}\selectfont
\begin{tabular}{llccccccc}
\toprule
\multicolumn{1}{l|}{\multirow{2}{*}{Model}}                                                                    & \multicolumn{1}{l|}{\multirow{2}{*}{LLM}} & \multicolumn{3}{c|}{General VQA}                                                                                   & \multicolumn{4}{c}{Text-oriented VQA}                                                                                         \\
\multicolumn{1}{l|}{}                                                                                          & \multicolumn{1}{l|}{}                     & GQA                           & ScienceQA                     & \multicolumn{1}{c|}{VizWiz}                        & TextVQA                       & ChartQA                       & DocVQA                        & AI2D                          \\
 \midrule
\multicolumn{9}{c}{\textit{Standard-resolution LMMs}}                                                                                                                                                                                                                                                                                                                                                             \\ \midrule
\multicolumn{1}{l|}{InstructBLIP \citep{DBLP:conf/nips/InstructBLIP}}                                                                                          & \multicolumn{1}{l|}{Vicuna-7B}                     & 49.2                          & 60.5                          & \multicolumn{1}{c|}{34.5}                          & -                             & -                             & -                             & -                             \\
\multicolumn{1}{l|}{IDEFICS-9B \citep{IDEFICS}}                                               & \multicolumn{1}{l|}{LLaMA-7B}             & 38.4                          & -                             & \multicolumn{1}{c|}{35.5}                          & 25.9                          & -                             & -                             & -                             \\
\multicolumn{1}{l|}{Qwen-VL \citep{qwen-vl}}                                                  & \multicolumn{1}{l|}{Qwen-7B}              & 59.3                          & 67.1                          & \multicolumn{1}{c|}{35.2}                          & 63.8                          & \underline{65.7} & 65.1                          & 62.3                          \\
\multicolumn{1}{l|}{Qwen-VL-Chat \citep{qwen-vl}}                                             & \multicolumn{1}{l|}{Qwen-7B}              & 57.5                          & 68.2                          & \multicolumn{1}{c|}{38.9}                          & 61.6                          & \textbf{66.3}                 & 62.6                          & 57.7                          \\
\multicolumn{1}{l|}{InternVL-Chat \citep{DBLP:journals/corr/InternVL-Chat}}                   & \multicolumn{1}{l|}{Vicuna-7B}            & 62.9                          & -                             & \multicolumn{1}{c|}{52.5}                          & 57.0                          & -                             & -                             & -                             \\
\multicolumn{1}{l|}{mPLUG-Owl2 \citep{DBLP:journals/corr/mPLUG-Owl2}}                         & \multicolumn{1}{l|}{LLaMA2-7B}            & 56.1                          & 68.7                          & \multicolumn{1}{c|}{54.5}                          & 58.2                          & -                             & -                             & -                             \\
\multicolumn{1}{l|}{MQT-LLAVA \citep{DBLP:journals/corr/MQT-LLAVA}}                           & \multicolumn{1}{l|}{Vicuna-7B}            & 61.6                          & 67.6                          & \multicolumn{1}{c|}{53.1}                          & -                             & -                             & -                             & -                             \\
\multicolumn{1}{l|}{LLaVA-1.5 \citep{Liu_2024_CVPR}}                                        & \multicolumn{1}{l|}{Vicuna-7B}            & 62.0                          & 66.8                          & \multicolumn{1}{c|}{50.0}                          & 58.2                          & -                             & -                             & -                             \\
\midrule
\multicolumn{9}{c}{\textit{High-resolution LMMs}}                                                                                                                                                                                                                                                                                                                                                               \\ \midrule
\multicolumn{1}{l|}{SPHINX-2k \citep{DBLP:journals/corr/SPHINX}}                              & \multicolumn{1}{l|}{LLaMA2-7B}            & \underline{63.1} & 70.6                          & \multicolumn{1}{c|}{44.9}                          & 61.2                          & -                             & -                             & -                             \\
\multicolumn{1}{l|}{TextMonkey \citep{DBLP:journals/corr/textmonkey}}                         & \multicolumn{1}{l|}{Qwen-VL-7B}           & -                             & -                             & \multicolumn{1}{c|}{-}                             & 65.9 & 58.2                          & 64.3                          & -                             \\
\multicolumn{1}{l|}{Mini-Gemini-HD \citep{DBLP:journals/corr/Mini-Gemini}}                         & \multicolumn{1}{l|}{Vicuna-7B}           & -                             & -                             & \multicolumn{1}{c|}{-}                             & \textbf{68.4} & -                          & -                          & -                             \\
\multicolumn{1}{l|}{MG-LLaVA \citep{DBLP:journals/corr/abs-2406-17770}}                                      & \multicolumn{1}{l|}{Vicuna-7B}            & 62.7                 & 70.4                          & \multicolumn{1}{c|}{\textbf{60.0}} & 58.4                          & 40.8                          & 44.6 & 64.1                          \\
\multicolumn{1}{l|}{LLaVA-NeXT \citep{liu2024llavanext}}                                      & \multicolumn{1}{l|}{Vicuna-7B}            & \textbf{64.2}                 & 70.1                          & \multicolumn{1}{c|}{57.6} & 64.9                          & 54.8                          & \underline{74.4} & 66.6                          \\
\multicolumn{1}{l|}{LLaVA-NeXT-\textit{M}$^3$ \citep{DBLP:journals/corr/m3}} & \multicolumn{1}{l|}{Vicuna-7B}            & -                             & \textbf{72.5}                 & \multicolumn{1}{c|}{-}                             & 63.1                          & 59.0                          & 72.6                          & \underline{66.7} \\ \midrule
\rowcolor[HTML]{EFEFEF} 
\multicolumn{1}{l|}{AVG-LLaVA}                                    & \multicolumn{1}{l|}{Vicuna-7B}            & 63.0                          & \underline{71.1} & \multicolumn{1}{c|}{\underline{59.8}}                 & \underline{67.1}                 & \textbf{66.3}                 & \textbf{74.6}                 & \textbf{67.3}                 \\ \bottomrule
\end{tabular}
\caption{Comparison with LMMs of the same size on general VQA benchmarks and text-oriented VQA benchmarks. The best results are marked in bold, and the second best results are underlined. Since MG-LLaVA is trained on significantly more data across two stages, we retrain it using the same data as ours for a fair comparison. We also explore the impact of additional two-stage training on the performance of LLaVA-NeXT using the same instruction fine-tuning data in Appendix \ref{sec:extra_training}.}
\label{tab:main_result_1}
\end{table*}

\paragraph{Training Datasets.} In the first training stage, since the real user interaction data used for visual instruction fine-tuning in LLaVA-NeXT are not open-sourced, we opt to extract 200K samples from the ALLaVA \citep{DBLP:journals/corr/ALLaVA} dataset as a substitute. Although LLaVA-NeXT replaces TextVQA \citep{DBLP:conf/cvpr/TextVQA} with DocVQA \citep{DBLP:conf/wacv/docvqa} and SynDog-EN \citep{DBLP:conf/eccv/SynDog-EN}, the TextVQA has already been included in the training data of most existing LMMs. Consequently, we choose to retain it to ensure a fair comparison with other models.\footnote{Our data recipe follows Open-LLaVA-NeXT \citep{Open-LLaVA-NeXT}.}
In total, the visual instruction fine-tuning data we use contains 1M image-text pairs.

\paragraph{Implementation Details.} Note that in this work, we focus on investigating the effectiveness of adaptive visual granularity selection in reducing the number of visual tokens and improving model performance, rather than building a state-of-the-art model.
Therefore, we use LLaVA-NeXT \cite{liu2024llavanext} as the base LMM, where the visual encoder is CLIP ViT-L/14, and the LLM is Vicuna-7B \citep{vicuna2023}. We set the filtered instruction token number $k$ to 32 and the cross-entropy loss weight $\alpha$ to 0.1.\footnote{The impact of these two hyperparameters on model performance is discussed in Appendix \ref{sec:hyper_analysis}.} In the first stage, the learning rates for the visual encoder and other modules are set to $2$$\times$10$^{-5}$ and $1$$\times$10$^{-5}$, respectively, with a batch size of 128. In the second stage, the learning rate for the visual granularity router is set to $1$$\times$$10^{-3}$, with a batch size of 128. More details of the training process are provided in Appendix \ref{sec:details}.

\paragraph{Evaluations.} We evaluate our model on three kinds of benchmarks: (1) \textbf{general VQA benchmarks}: GQA \citep{DBLP:conf/cvpr/GQA}, SciQA-Img \citep{DBLP:conf/nips/sqa}, and VizWiz \citep{DBLP:conf/cvpr/VizWiz}; (2) \textbf{text-oriented VQA benchmarks}: TextVQA \citep{DBLP:conf/cvpr/TextVQA}, ChartQA \citep{DBLP:conf/acl/chartqa}, DocVQA \citep{DBLP:conf/wacv/docvqa},  and AI2D \citep{DBLP:conf/eccv/ai2d}; and (3) \textbf{general multimodal benchmarks}: MME \citep{DBLP:journals/corr/mme}, MMB \citep{DBLP:journals/corr/mmb}, MMB$^{CN}$ \citep{DBLP:journals/corr/mmb}, POPE \citep{DBLP:conf/emnlp/POPE}, and MMMU \citep{DBLP:journals/corr/MMMU}.

\subsection{Main Results}

\begin{table*}[t!]
\setlength\tabcolsep{9.5pt}
\centering
\renewcommand\arraystretch{1.2}
\small
\fontsize{8pt}{8pt}\selectfont
\begin{tabular}{llcccccc}
\toprule
\multicolumn{1}{c|}{Model}                                                                                     & \multicolumn{1}{c|}{LLM}       & MME                           & MME$^{C}$                          & MMB                           & MMB$^{CN}$                    & POPE                          & MMMU                          \\ \midrule
\multicolumn{8}{c}{\textit{Standard-resolution LMMs}}                                                                                                                                                                                                                                                                                                           \\ \midrule
\multicolumn{1}{l|}{InstructBLIP \citep{DBLP:conf/nips/InstructBLIP}}                         & \multicolumn{1}{l|}{Vicuna-7B} & 1084.0                            & 229.0                            & -                             & -                             & -                             & 30.6                          \\
\multicolumn{1}{l|}{Qwen-VL-Chat \citep{qwen-vl}}                                             & \multicolumn{1}{l|}{Qwen-7B}   & 1487.6                          & \underline{360.7}                          & 60.6                          & -                             & -                             & -                             \\
\multicolumn{1}{l|}{InternVL-Chat \citep{DBLP:journals/corr/InternVL-Chat}}                   & \multicolumn{1}{l|}{Vicuna-7B} & 1525.1                          & -                              & -                             & -                             & 86.4                          & -                             \\ 
\multicolumn{1}{l|}{mPLUG-Owl2 \citep{DBLP:journals/corr/mPLUG-Owl2}}                         & \multicolumn{1}{l|}{LLaMA2-7B} & 1450.2                          & -                              & 64.5                          & -                             & -                             & -                             \\
\multicolumn{1}{l|}{MQT-LLAVA \citep{DBLP:journals/corr/MQT-LLAVA}}                           & \multicolumn{1}{l|}{Vicuna-7B} & 1434.5                          & 353.6 & 64.3                          & -                             & 84.4                          & 34.8                          \\
\multicolumn{1}{l|}{LLaVA-1.5 \citep{DBLP:conf/nips/LiuLWL23a}}                               & \multicolumn{1}{l|}{Vicuna-7B} & 1510.7                          & -                              & 64.3                          & 58.3                          & \underline{87.3}                          & -                             \\
\midrule
\multicolumn{8}{c}{\textit{High-resolution LMMs}}                                                                                                                                                                                                                                                                                                             \\ \midrule
\multicolumn{1}{l|}{SPHINX-2k \citep{DBLP:journals/corr/SPHINX}}                              & \multicolumn{1}{l|}{LLaMA2-7B} & 1470.6                          & 326.8                          & 65.9                          & -                             & 87.2 & -                             \\
\multicolumn{1}{l|}{OtterHD-8B \citep{DBLP:journals/corr/Otterhd}}                            & \multicolumn{1}{l|}{Fuyu-8B}   & 1223.4                          & 331.4                          & 58.3                          & -                             & 86.0                          & -                             \\
\multicolumn{1}{l|}{Mini-Gemini-HD \citep{DBLP:journals/corr/Mini-Gemini}}                            & \multicolumn{1}{l|}{Vicuna-7B}   & 1546.0                          & 319.0                          & 65.8                          & -                             & -                          & \underline{36.8}                             \\
\multicolumn{1}{l|}{MG-LLaVA \citep{DBLP:journals/corr/abs-2406-17770}}                                      & \multicolumn{1}{l|}{Vicuna-7B} & \textbf{1561.1} & 325.4                            & 67.4                          & 48.4 & 86.9                          & 35.3 \\
\multicolumn{1}{l|}{LLaVA-NeXT \citep{liu2024llavanext}}                                      & \multicolumn{1}{l|}{Vicuna-7B} & 1519.0 & 332.0                            & 67.4                          & \underline{60.6} & 86.5                          & 35.8 \\
\multicolumn{1}{l|}{LLaVA-NeXT-\textit{M}$^3$ \citep{DBLP:journals/corr/m3}} & \multicolumn{1}{l|}{Vicuna-7B} & -                               & -                              & \underline{68.0} & -                             & 87.2 & 34.0                          \\ \midrule
\rowcolor[HTML]{EFEFEF}
\multicolumn{1}{l|}{AVG-LLaVA}                                    & \multicolumn{1}{l|}{Vicuna-7B} & \underline{1557.4}                 & \textbf{366.8}                 & \textbf{69.9}                 & \textbf{61.8}                 & \textbf{87.4}                 & \textbf{37.4}                 \\ \bottomrule
\end{tabular}
\caption{Comparison with LMMs of the same size on general multimodal benchmarks.}
\label{tab:main_result_2}
\end{table*}

\paragraph{General VQA Benchmarks.} The results in Table \ref{tab:main_result_1} show that AVG-LLaVA outperforms all standard-resolution LMMs on the general VQA benchmarks and achieves comparable performance to other high-resolution LMMs. Although it does not achieve the best results, it is important to note that AVG-LLaVA uses fewer visual tokens compared to other high-resolution models, and this comparison will be detailed in Section \ref{sec:efficiency}.
\paragraph{Text-oriented VQA Benchmarks.} In this category of benchmarks, as illustrated in Table \ref{tab:main_result_1}, except for TextVQA, AVG-LLaVA outperforms all other comparison models. Back to TextVQA, AVG-LLaVA achieves the second-best performance, only trailing behind Mini-Gemini-HD. Notably, Mini-Gemini-HD utilizes more than twice the amount of data during the pretraining and approximately 1.5 times the amount of data during the visual instruction fine-tuning compared to AVG-LLaVA.

\begin{table*}[ht!]
\centering
\setlength\tabcolsep{8pt}
\renewcommand\arraystretch{1.2}
\small
\fontsize{8pt}{8pt}\selectfont
\begin{tabular}{l|ccc|ccc|ccc}
\toprule
\multirow{2}{*}{Metric} & \multicolumn{3}{c|}{General VQA} & \multicolumn{3}{c|}{Text-oriented VQA} & \multicolumn{3}{c}{MLLM Benchmarks} \\
                        & GQA      & ScienceQA   & VizWiz  & TextVQA     & ChartQA     & AI2D       & MME        & MMB        & MMMU      \\ \midrule
Token Per Grid $\downarrow$         & 80.0\%   & 26.4\%      & 54.9\%  & 92.3\%      & 99.1\%      & 14.7\%     & 69.3\%     & 30.0\%     & 29.9\%    \\
Speed $\uparrow$                  & 1.14$\times$     & 1.77$\times$       & 1.41$\times$   & 1.04$\times$       & 0.97$\times$       & 2.53$\times$      & 1.19$\times$      & 1.87$\times$      & 1.79$\times$  
\\ \bottomrule
\end{tabular}
\caption{Comparisons of AVG-LLaVA and LLaVA-NeXT in terms of the number of visual tokens and actual inference speed, both of which are tested on 8 V100 GPUs with a batch size of 1. AVG-LLaVA can reduce the number of visual tokens by up to 85.3\% and is up to 2.53$\times$ faster than LLaVA-NeXT.}
\label{tab:speed}
\end{table*}

\paragraph{General Multimodal Benchmarks.} Compared to traditional VQA datasets, this type of benchmarks cover a broader range of evaluation aspects, requiring models to possess more complex perception and reasoning capabilities. As summarized in Table \ref{tab:main_result_2}, except for MME, AVG-LLaVA surpasses all other models across the remaining benchmarks, exhibiting superior overall performance and highlighting its adaptability and effectiveness across various disciplines. Specifically, AVG-LLaVA outperforms the second-best model by 6.1, 1.9, and 1.2 on MME$^{C}$, MMB, and MMB$^{CN}$, respectively. Moreover, AVG-LLaVA's performance on the POPE and MMMU benchmarks demonstrates its ability to reduce hallucinations and perform complex reasoning.

\subsection{Computational Efficiency}
\label{sec:efficiency}
To validate the effectiveness of dynamic visual granularity selection, we compare AVG-LLaVA with LLaVA-NeXT in terms of visual token number and inference speed across multiple benchmarks. Specifically, for each type of benchmarks, we select three benchmarks for comparison, and report the reduction in the number of visual tokens per grid and the actual speedup during inference. 

As shown in Table \ref{tab:speed}, except for text-intensive VQA benchmarks that require very fine-grained visual information, such as TextVQA and ChartVQA, AVG-LLaVA significantly reduces the number of visual tokens and improves inference speed across other benchmarks. Particularly, on the AI2D benchmark, AVG-LLaVA achieves better performance than LLaVA-NeXT while using only 14.7\% of visual tokens, and the inference speed increases by 2.53 $\times$.\footnote{We also present qualitative results in Appendix \ref{sec:qualitative} and illustrate the effectiveness of adaptive visual granularity.} Notably, even with the addition of two extra modules, there is no significant slowdown in inference speed on the ChartVQA benchmark when using a comparable number of visual tokens. Moreover, AVG-LLaVA only increases the number of parameters by 1.66\% compared to LLaVA-NeXT.

\subsection{Routing Visualization}
\begin{figure*}[t]
    \centering
    \includegraphics[width=0.95\linewidth]{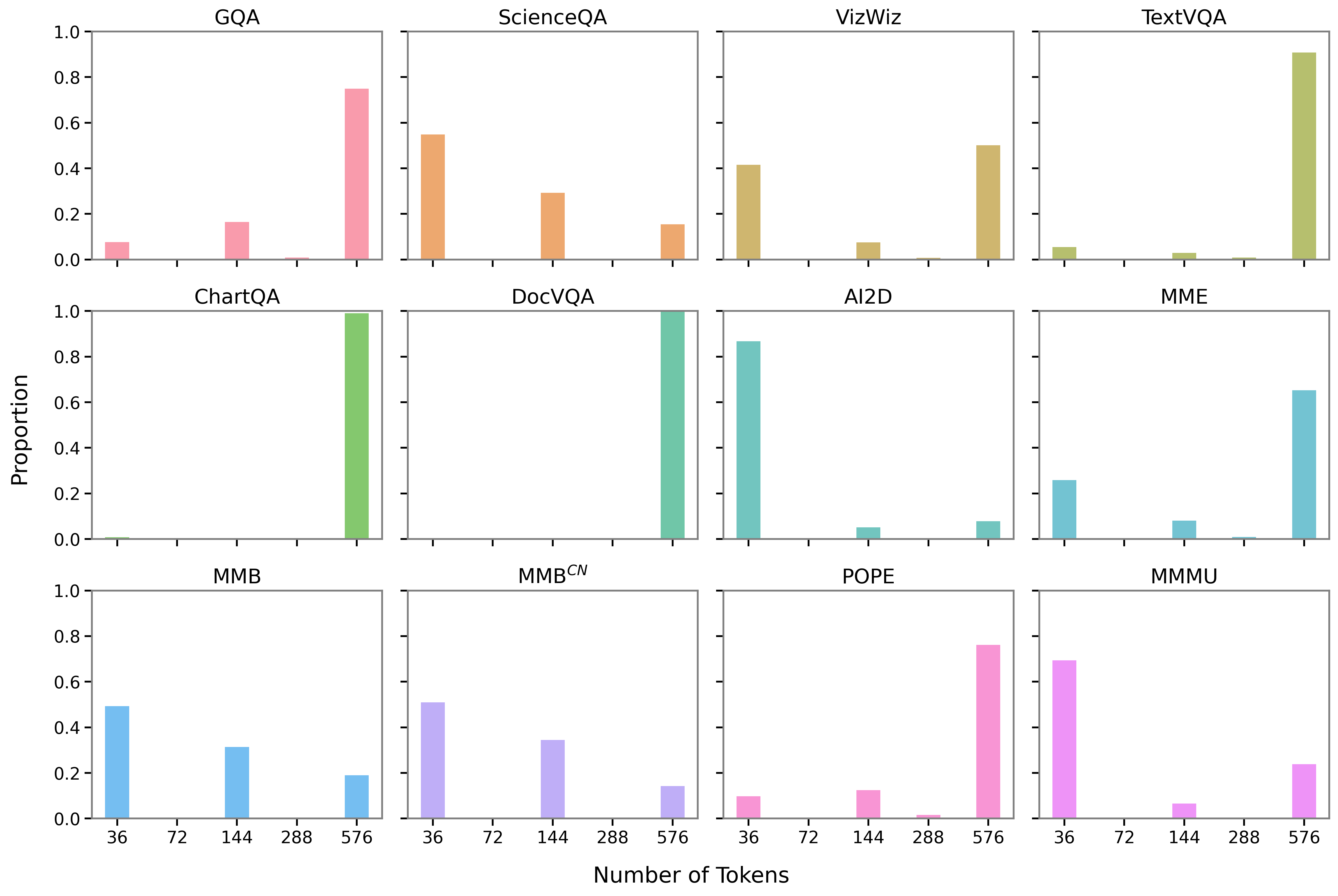}
    \vspace{-3mm}
    \caption{Visualization of the proportion for different granularity visual tokens.}
    \label{fig:visualization}
\end{figure*}

\label{sec:visualize}
To further understand the differences in the granularity selection of AVG-LLaVA across different benchmarks, we visualize the proportion of visual tokens selected at each granularity level for all benchmarks. Figure \ref{fig:visualization} shows the visualization results, it is evident that different tasks tend to favor different visual granularity, which is consistent with our expectations. In the case of text-intensive benchmarks like TextVQA, ChartQA, and DocVQA, the model requires fine-grained visual information, so the router predominantly selects the finest visual granularity. On the other hand, for benchmarks involving object-level questions, such as AI2D and MMMU, the model may find it easier to answer correctly by utilizing coarse-grained visual information. Although the 72 and 288-token granularities are seldom selected, their inclusion helps the model progressively learn and differentiate between various levels of visual granularity (see the ablation study in Section \ref{sec:ablation}).

\subsection{Ablation Study}
\begin{table*}[ht!]
\resizebox{\textwidth}{!}{
\begin{tabular}{lccccc|cclc}
\toprule
Ablated Setting               &     & Ablated Details       & \begin{tabular}[c]{@{}c@{}}Original Value\end{tabular} & → & Changed Value                                                             & ScienceQA & ChartQA & MME    & MMB  \\ \midrule
\multicolumn{6}{c|}{\textbf{AVG-LLaVA}}                                                                                                                                                                                     & 71.1      & 66.3    & 1557.4 & 69.9 \\ \midrule
\multirow{4}{*}{Architecture} & (a) & Visual Granularity & Adaptive                                                     &   & Fixed                                                          & 70.0      & 66.4    & 1554.5 & 68.7 \\
& (b) & Granularity Selection & Router                                                     &   & Random                                                          & 69.7      & 56.8    & 1535.7 & 67.9 \\
                              & (c) & Router Input          & Image + Instruction                                                  &   & Image                                                                     & 70.1      & 53.9    & 1525.2 & 69.0 \\
                              & (d) & Granularity Range    & \{36, 72, 144, 288, 576\}                                            &   & \{36, 144, 576\}                                                          & 69.8      & 65.3    & 1547.7 & 66.3 \\ \midrule
\multirow{3}{*}{Training}     & (e) & Router Training       & Feedback from LMM                                                    &   & \begin{tabular}[c]{@{}c@{}}Visual Instruction \\ Fine-tuning\end{tabular} &70.5           &50.9         &1514.8        &68.6      \\
                              & (f) & Ranking Loss          & \Checkmark                                            &   & \XSolidBrush                                               & 70.1      & 64.8    & 1534.6 & 68.6 \\
                              & (g) & Cross-entropy Loss    & \Checkmark                                            &   & \XSolidBrush                                               & 70.2      & 66.3    & 1550.8 & 69.4 \\ \bottomrule
\end{tabular}
}
\caption{Ablation results on multiple benchmarks.}
\label{tab:ablation}
\end{table*}

\label{sec:ablation}
In order to validate the effectiveness of our designed modules and training paradigm, we conduct the following ablation analysis.

\begin{figure*}[t]
    \centering
    \includegraphics[width=1\linewidth]{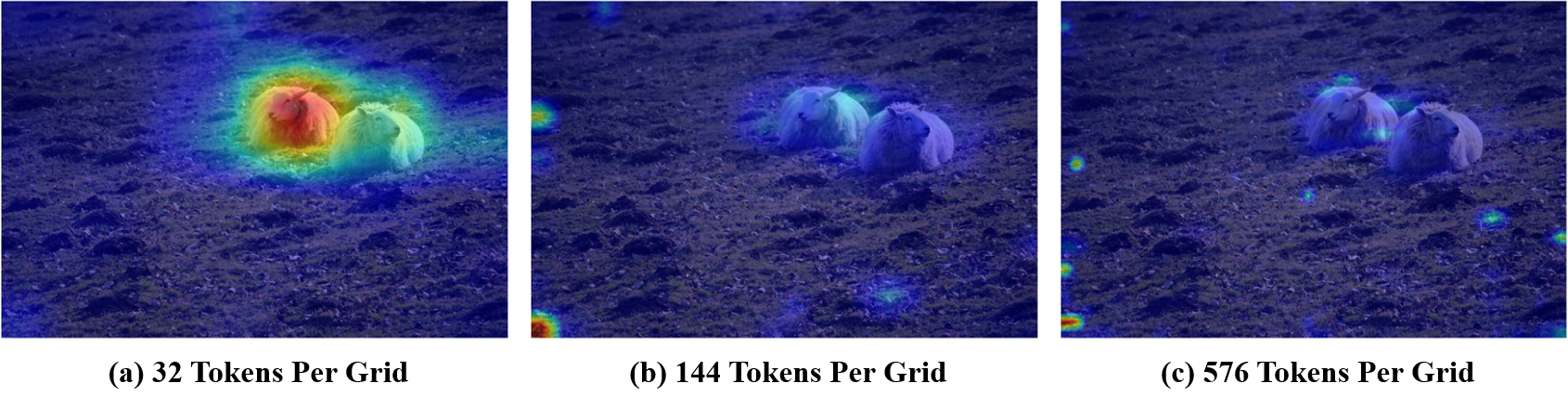}
    \vspace{-7mm}
    \caption{Attention maps of AVG-LLaVA at different visual granularities. The instruction is ``\textit{How many sheep are there? Answer the question with a single word.}''. The number of visual tokens per grid selected by the router is 32.}
    \label{fig:attention_map}
\end{figure*}

\paragraph{Adaptive Visual Granularity vs. Fixed Visual Granularity.} We first delve into the proposed adaptive visual granularity router and report results in Table \ref{tab:ablation}(a). It is clear that, compared to fixed visual granularity, adaptive visual granularity shows significant improvement on ScienceQA, MME, and MMB. It is worth noting that, in addition to performance improvement, adaptive visual granularity can also significantly reduce the number of visual tokens and increase the model's inference speed, as reported in Section \ref{sec:efficiency}.

\paragraph{Router Granularity Selection vs. Random Granularity Selection.} In Table \ref{tab:ablation}(b), we replace the granularity selected by the router with randomly-selected granularity. The results show that visual granularity router can indeed select a relatively appropriate granularity based on the input image and instruction, thereby significantly enhancing model performance.

\paragraph{Impact of Router Input.} The instruction plays a crucial role in granularity selection. To validate this, we remove the instruction from the router input. As shown in Table \ref{tab:ablation}(c), a clear performance degradation rises when solely using image as input (e.g, -12.4 on ChartQA), illustrating the importance of choosing granularity based on input image and instruction.

\paragraph{Impact of Granularity Range.} In Section \ref{sec:visualize}, we observe that granularities with 72 and 288 visual tokens are rarely selected, therefore we remove the visual tokens of these two granularities. As shown in Table \ref{tab:ablation}(d), this change leads to a decrease in model performance, proving that introducing these granularities benefits the model's progressive learning to utilize features of different visual granularities and distinguish among various visual granularities.

\paragraph{Impact of Router Training Methods.} We directly train the router using visual instructions fine-tuning with the cross-entropy loss function. Unlike our original approach where the router is directly supervised by the LMM feedback, this variant computes the loss on the LMM and backpropagates the gradient to the router using the Gumbel-Softmax technique \citep{DBLP:conf/iclr/gumbel-softmax}. The results in Table \ref{tab:ablation}(e) show that the LLM feedback allows the router to better distinguish the advantages and disadvantages of different granularities, thereby enabling it to select an appropriate granularity.

\paragraph{Importance of Ranking Granularity.} In Table \ref{tab:ablation}(f) and Table \ref{tab:ablation}(g), we remove the cross-entropy loss and ranking loss during the second stage, respectively. The results indicate that both types of loss are beneficial to model training and are complementary to each other, between which the ranking loss is more crucial. This underscores the necessity to train the router by ranking granularity based on LMM feedback.

\subsection{Attention Map Visualization}
To further understand how the appropriate granularity benefits the model in generating better answers, we visualize the attention map between the generated tokens and the visual tokens. The attention weights are calculated by accumulating the attention scores between image tokens and generated tokens across all layers and heads. As shown in Figure \ref{fig:attention_map},  when the instruction is ``\textit{How many sheep are there? Answer the question with a single word.}'' the attention weights for the visual granularity selected by the router are mostly assigned to the two sheep, while the attention weights for other visual granularities are dispersed across the background. This means that selecting the appropriate visual granularity results in a more distinct attention map characterized by reduced background noise and enhanced focus on relevant regions, thereby improving model performance. 
\section{Conclusion}
In this work, we propose AVG-LLaVA, an LMM that can adaptively select appropriate visual granularity based on input image and instruction. 
Besides, we introduce RGLF, which aligns router-predicted probabilities of multiple granularities with LMM preferences by a ranking loss, effectively helping the model learn to distinguish between different granularities. Experimental results show that AVG-LLaVA not only exhibits superior performance across 11 benchmarks, but also significantly reduce the number of visual tokens and speed up inference in tasks that do not require fine-grained information. In future work, we aim to develop different visual granularity scaling networks to obtain richer visual granularity and integrate the two-stage training into a single stage to improve efficiency. 

\section*{Limitations}
While AVG-LLaVA has achieved good results, there is still considerable potential to be further explored. On text-intensive benchmarks, the model tends to select the finest-grained visual tokens, which may be due to the pooling directly reducing half of the tokens, resulting in significant differences in granularity size. Designing  a more suitable granularity scaling network to provide richer visual granularities may help alleviate this issue. Besides, the two-stage training introduces additional overhead, which could be alleviated by interleaving multi-granularity visual instruction fine-tuning and router training within a single stage.

\section*{Acknowledgments}
The project was supported by 
National Key R\&D Program of China (No. 2022ZD0160501), 
Natural Science Foundation of Fujian Province of China (No. 2024J011001),
and
the Public Technology Service Platform Project of Xiamen (No.3502Z20231043).
We also thank the reviewers for their insightful comments.

\bibliography{custom}

\newpage
\appendix

\begin{algorithm*}[t]
\caption{Visual Granularity Selection Algorithm}
\begin{algorithmic}[1]
\REQUIRE Multi-granularity visual tokens $\mathbf{\overline{X}}_v = [\mathbf{X}_{v}^1;\mathbf{X}_{v}^2;...;\mathbf{X}_{v}^N]$, Instruction tokens $\mathbf{X}_{instruct}$, \\
Visual tokens of original granularity $\mathbf{X}_v$

\ENSURE Selected granularity visual tokens $\mathbf{X}_v^{selected}$.

\STATE Obtain the filtered instruction tokens $\mathbf{\overline{X}}_{instruct}$ = Top-$k$(cosine\_sim($\mathbf{X}_{instruct},\mathbf{X}_v$))

\STATE Concatenate $\mathbf{\overline{X}}_v$ and $\mathbf{\overline{X}}_{instruct}$ to form the input for the router

\STATE Apply a Transformer layer to facilitate token fusion $\mathbf{Z}_{fusion} = \text{Transformer}([\mathbf{\overline{X}}_v; \mathbf{\overline{X}}_{instruct}])$.

\STATE Use an MLP to predict logits for each token $\mathbf{Z}_{out} = \text{MLP} (\mathbf{Z}_{fusion})$

\STATE Aggregate the logits using a learnable weight matrix $\mathbf{Z}_{final} = \mathbf{W} \mathbf{Z}_{out}$.

\STATE Compute the probability distribution using softmax $\mathbf{P} = \text{softmax}(\mathbf{Z}_{final})$. 

\STATE Identify the granularity with the highest probability ${selected} = \arg\max(\mathbf{P})$.

\RETURN $\mathbf{X}_v^{selected}$

\end{algorithmic}
\label{algorithm:router}
\end{algorithm*}

\section{Appendix}
\subsection{Visual Granularity Selection Algorithm}
In algorithm \ref{algorithm:router}, we provide the detailed process of the router's granularity selection.

\subsection{Hyperparameter Analysis}
\label{sec:hyper_analysis}
\begin{figure}[h]
    \centering
    \begin{minipage}[b]{1\linewidth}
        \centering
        \setlength{\abovecaptionskip}{-2pt}
        \includegraphics[width=\linewidth]{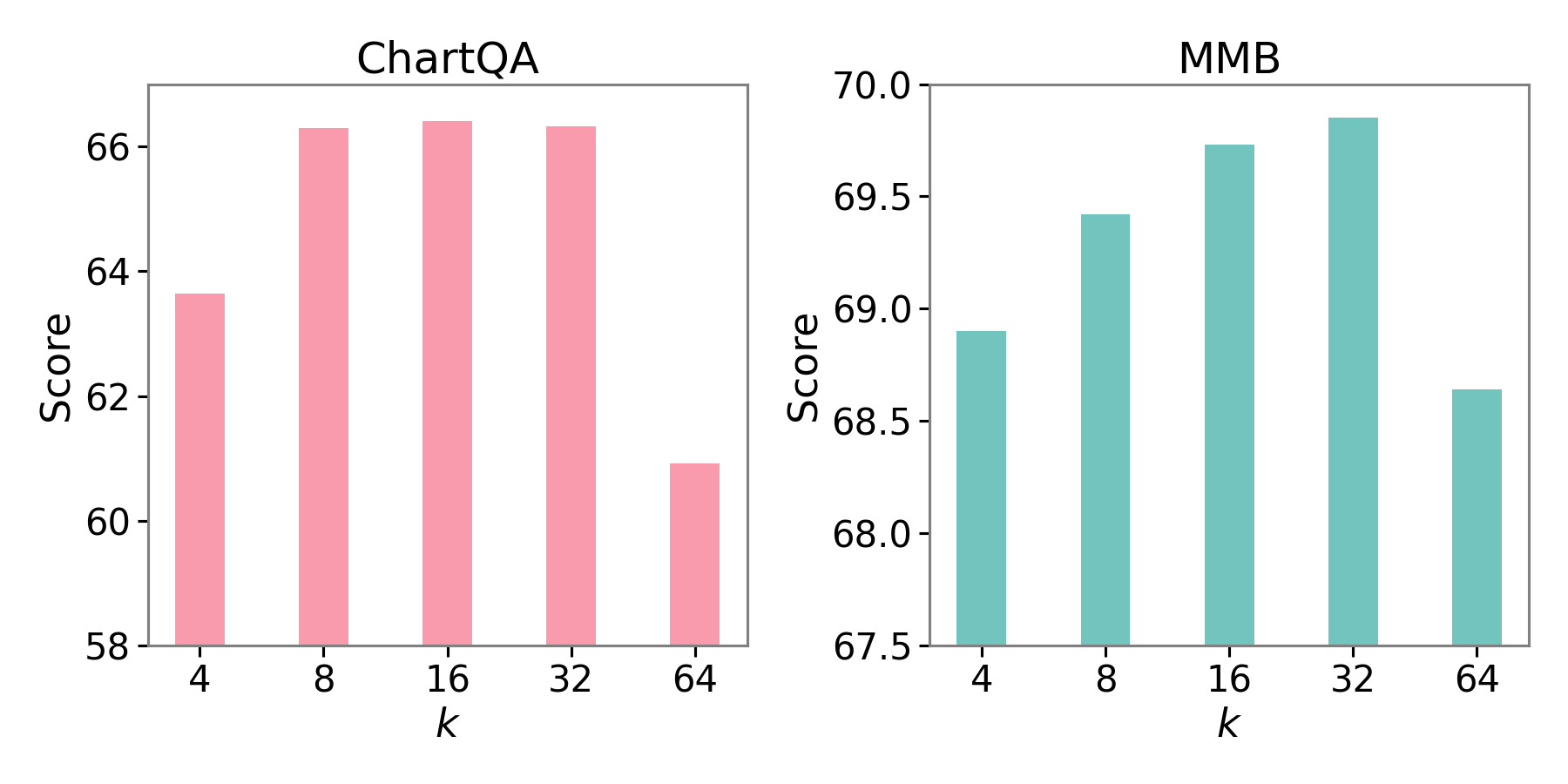}
        \caption{Influence of the filtered instruction token number $k$ on model performance, measured on ChartQA and MMB benchmarks.}
        \label{fig:k}
    \end{minipage}
    \hfill
    \begin{minipage}[b]{1\linewidth}
        \centering
        \setlength{\abovecaptionskip}{-2pt}
        \includegraphics[width=\linewidth]{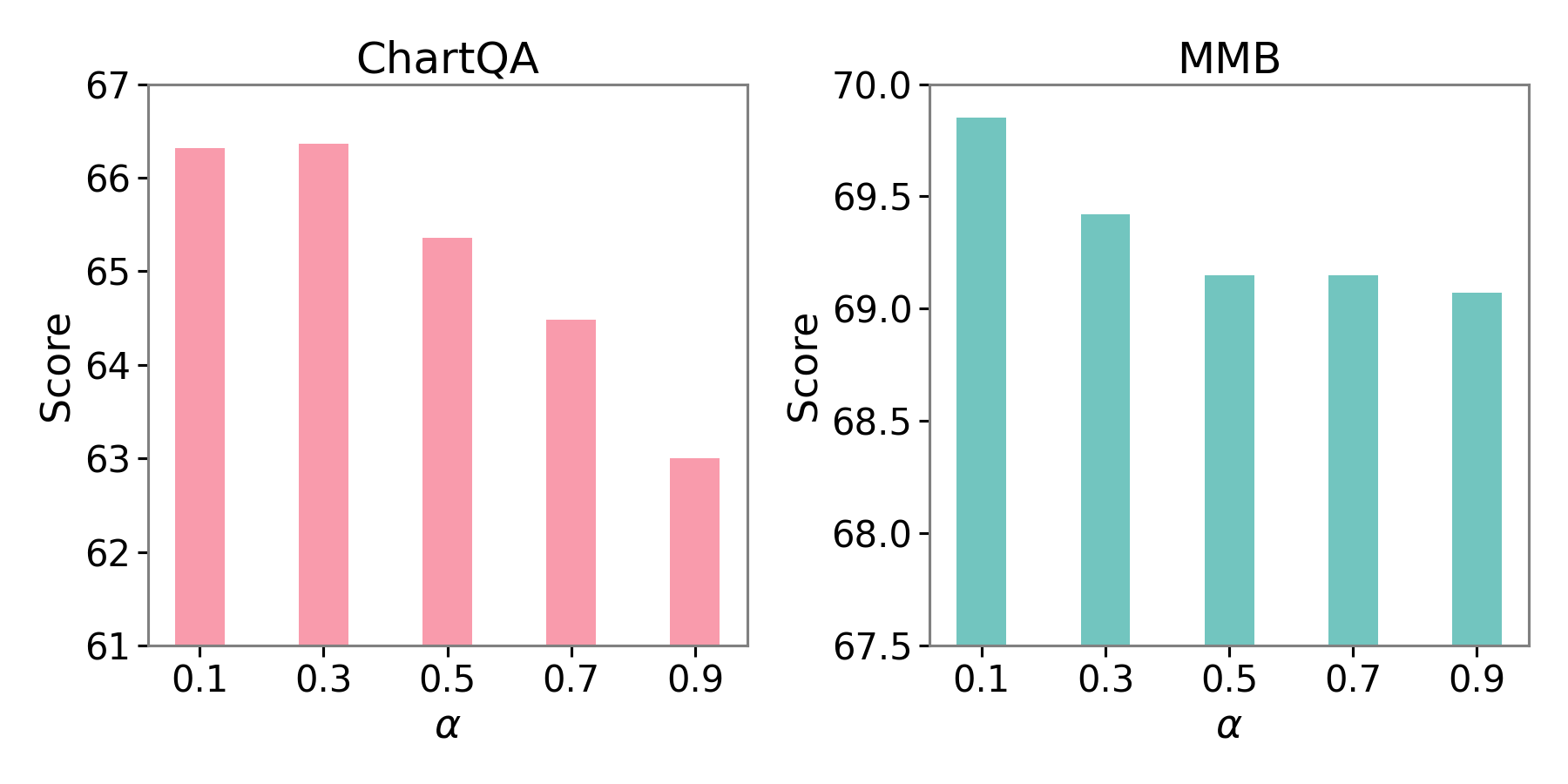}
        \caption{Influence of the cross-entropy loss weight $\alpha$ on model performance, measured on ChartQA and MMB benchmarks.}
        \label{fig:alpha}
    \end{minipage}
\end{figure}
We experimentally explore the influence of the filtered instruction token number $k$ and the cross-entropy loss weight $\alpha$ on model performance. As shown in Figure \ref{fig:k}, the model performance is significantly affected when $k$ is too small or too large. This may be due to the fact that too few instruction tokens provide insufficient text information, while too many tokens will introduce more noise. Figure \ref{fig:alpha} indicates that our approach is relatively robust to $\alpha$ and setting a smaller $\alpha$ is able to consistently enhance model performance, making our training method easy to apply.

\subsection{Training Details}
\label{sec:details}
We list the training hyperparameters for
two stages in Table \ref{tab:hyper}. Our setup mainly refers to LLaVA-NeXT \cite{liu2024llavanext}.
\begin{table}[t]
\centering
\setlength\tabcolsep{8pt}
\renewcommand\arraystretch{1.2}
\small
\fontsize{9pt}{9pt}\selectfont
\begin{tabular}{l|cc}
\toprule
Hyperparameter    & Stage 1 & Stage 2 \\ \midrule
Data size        & 1M     & 1M     \\
Batch size        & 128     & 128     \\
lr                & 1e-5    & 1e-3    \\
Vision encoder lr & 2e-5    & -       \\
lr schedule       & \multicolumn{2}{c}{cosine decay}      \\
lr warmup ratio   & \multicolumn{2}{c}{0.03}              \\
Weight decay      & \multicolumn{2}{c}{0}                 \\
Epoch             & \multicolumn{2}{c}{1}                 \\
Optimizer         & \multicolumn{2}{c}{AdamW}             \\
DeepSpeed stage  & 3       & 3       \\
GPU  & 8 $\times$ H800       & 8 $\times$ H800       \\
Training cost (\#Hours)   & 65       & 14       \\
\bottomrule
\end{tabular}
\caption{Training hyperparameters of AVG-LLaVA. }
\label{tab:hyper}
\end{table}

\begin{table*}[]
\centering
\setlength\tabcolsep{8pt}
\renewcommand\arraystretch{1.2}
\small
\fontsize{9pt}{9pt}\selectfont
\begin{tabular}{l|ccc|cccc}
\toprule
\multirow{2}{*}{Model} & \multicolumn{3}{c|}{General VQA}               & \multicolumn{4}{c}{Text-oriented VQA}                         \\
                       & GQA           & ScienceQA     & VizWiz        & TextVQA       & ChartQA       & DocVQA        & AI2D          \\ \midrule
LLaVA-NeXT             & 64.2          & 70.1          & 57.6          & 64.9          & 54.8          & 74.4          & 66.6          \\
LLaVA-NeXT-Extra       & \textbf{64.6} & 69.9          & 58.3          & 63.9          & 66.3          & \textbf{75.1} & 65.3          \\
AVG-LLaVA              & 63.0          & \textbf{71.1} & \textbf{59.8} & \textbf{67.1} & \textbf{66.3} & 74.6          & \textbf{67.3} \\ \bottomrule
\end{tabular}
\caption{Results on general VQA benchmarks and text-oriented VQA benchmarks. LLaVA-NeXT-Extra refers to training for two extra epochs on the same multimodal instruction-tuning data.}
\label{tab:extra_vqa}
\end{table*}

\begin{table*}[t]
\centering
\setlength\tabcolsep{10pt}
\renewcommand\arraystretch{1.2}
\small
\fontsize{9pt}{9pt}\selectfont
\begin{tabular}{l|cccccc}
\toprule
Model            & MME             & MME$^{C}$      & MMB           & MMB$^{CN}$    & POPE          & MMMU          \\ \midrule
LLaVA-NeXT       & 1519.0          & 332.0          & 67.4          & 60.6          & 86.5          & 35.8          \\
LLaVA-NeXT-Extra & 1524.7          & 330.0          & 67.8          & 57.0          & \textbf{87.4} & 34.8          \\
AVG-LLaVA        & \textbf{1557.4} & \textbf{366.8} & \textbf{69.9} & \textbf{61.8} & \textbf{87.4} & \textbf{37.4} \\ \bottomrule
\end{tabular}
\caption{Results on general multimodal benchmarks.}
\label{tab:extra_general_multimodal}
\end{table*}
\subsection{Impact of Multiple Training Epochs on the Same Dataset}
\label{sec:extra_training}
Since we conduct two additional training stages on the LLaVA-NeXT, we also investigate the impact of training LLaVA-NeXT for two extra epochs on the same multimodal instruction-tuning data. 

From Tables \ref{tab:extra_vqa} and \ref{tab:extra_general_multimodal}, we can observe that although three additional two epoch trainings result in improvements on 7 benchmarks (e.g., ChartQA and DocVQA), there is a considerable performance decline on 6 benchmarks (e.g., TextVQA and MMB$^{CN}$). This indicates that repeated training cannot improve the performance on all benchmarks. Besides, AVG-LLaVA performs better than LLaVA-NeXT-Extra on 9 benchmarks, is slightly worse on 2 benchmarks, and has a significant speed improvement, indicating that the advantage of AVG-LLaVA does not simply stem from repeated training.

\subsection{Qualitative Evaluation}
\label{sec:qualitative}
\begin{figure*}[t]
    \centering
    \includegraphics[width=1\linewidth]{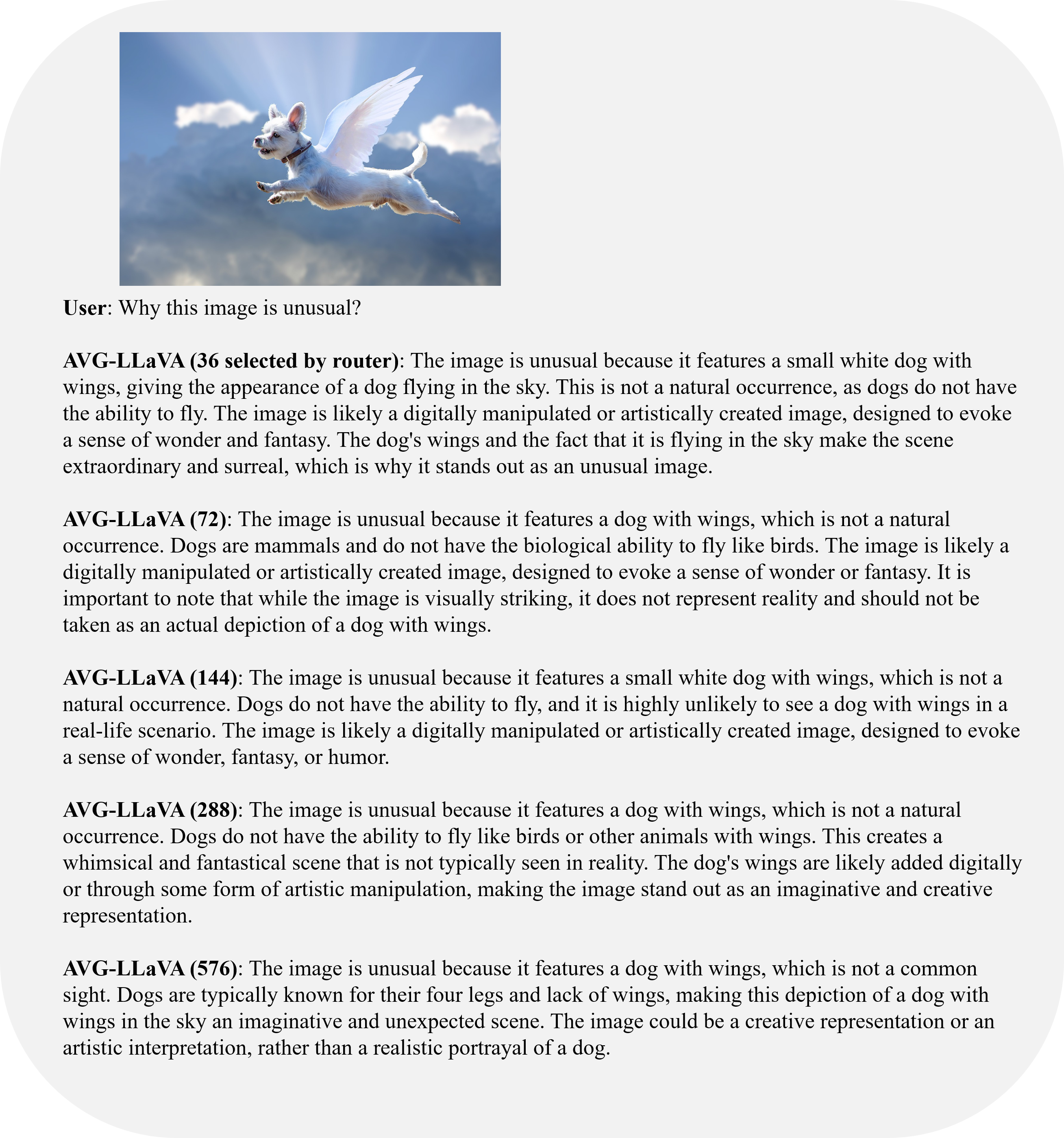}
    \caption{Example of the finest visual granularity selected by the router compared to other visual granularities. The number in parentheses indicates the number of visual tokens. In this example, the responses from the model using both fine-grained and coarse-grained visual tokens show little difference, indicating that the model can adopt coarse visual granularity to significantly reduce the number of visual tokens and accelerate inference.}
    \label{fig:case:36}
\end{figure*}

\begin{figure*}[t]
    \centering
    \includegraphics[width=1\linewidth]{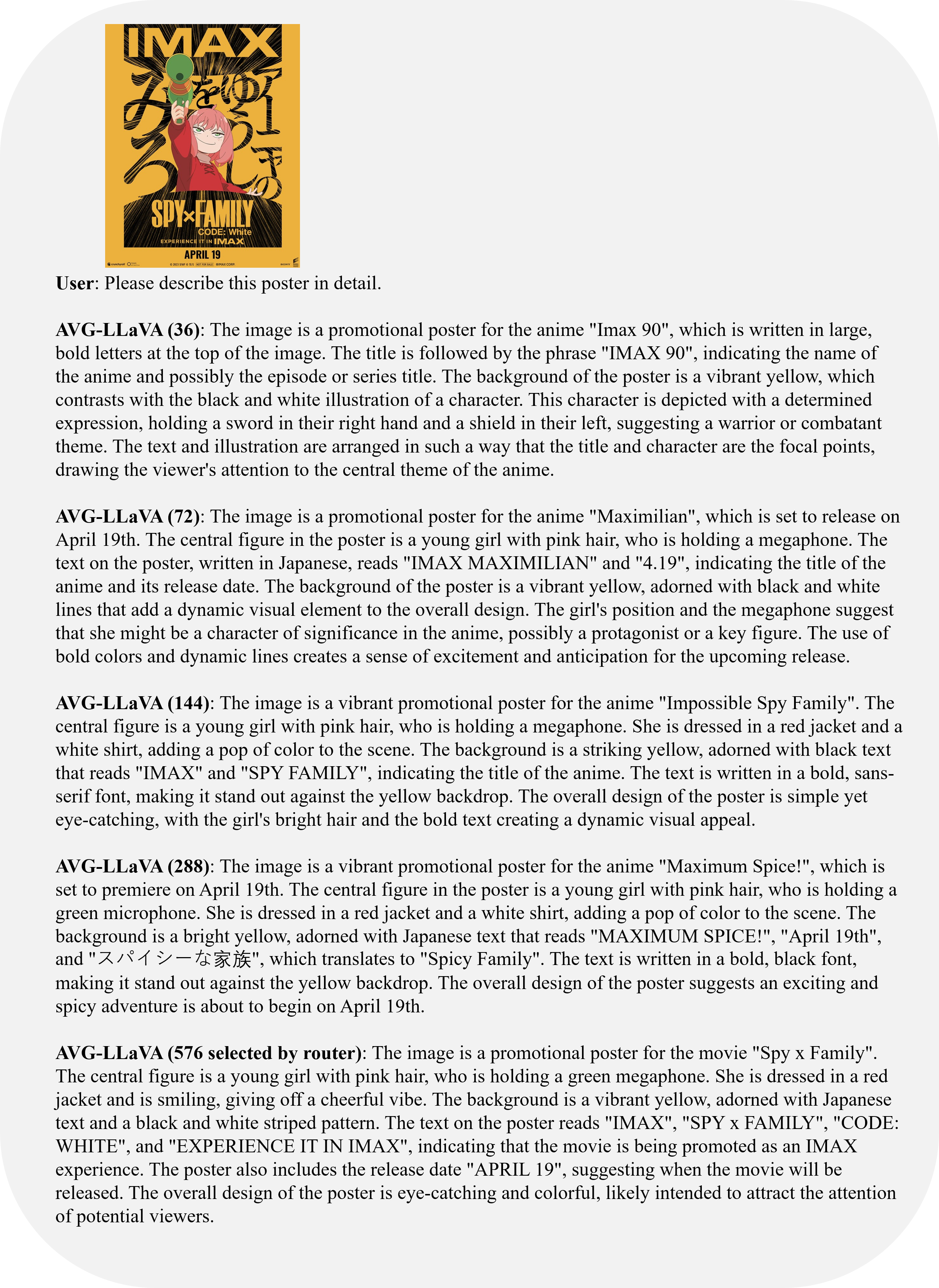}
    \caption{Example of the coarsest visual granularity selected by the router compared to other visual granularities. The number in parentheses indicates the number of visual tokens. In this example, when coarse-grained visual tokens are used, the model generates incorrect descriptions. This suggests that the model should select fine visual granularity for the image and instructions in order to achieve better accuracy.}
    \label{fig:case:576}
\end{figure*}

\begin{figure*}[t]
    \centering
    \includegraphics[width=1\linewidth]{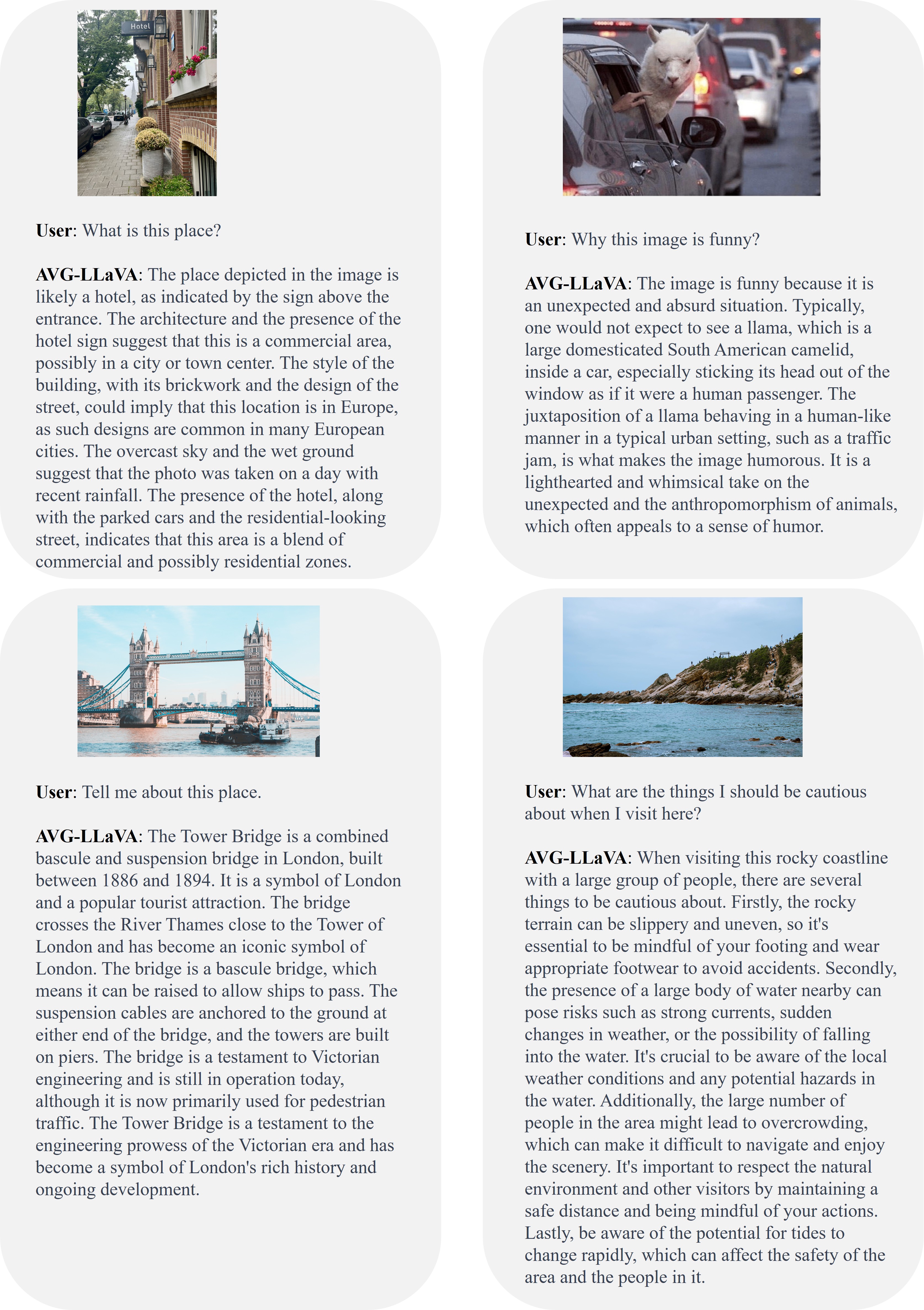}
    \caption{More Examples of conversations between users and AVG-LLaVA.}
    \label{fig:more_case}
\end{figure*}

As shown in Figures \ref{fig:case:36} and \ref{fig:case:576}, we compare the visual granularity selected by the router and other visual granularities. In Figure \ref{fig:case:36}, given the user-provided image and instruction, AVG-LLaVA selects the coarsest visual granularity through the router. It can be observed that compared to other granularities, the model's response with the coarsest granularity does not vary significantly. However, in Figure \ref{fig:case:576}, with the given image and instruction, AVG-LLaVA selects the finest visual granularity. We find that coarser visual granularities could not generate a reasonably accurate poster description. These two examples demonstrate that AVG-LLaVA can adaptively select the appropriate visual granularity based on the image and instruction, thereby reducing the number of visual tokens, accelerating inference, and even improving model performance.

Figure \ref{fig:more_case} further shows several conversations between users and AVG-LLaVA.

\subsection{Visualization of Router Selection for Different Instructions}
\begin{figure*}[t]
    \centering
    \includegraphics[width=1\linewidth]{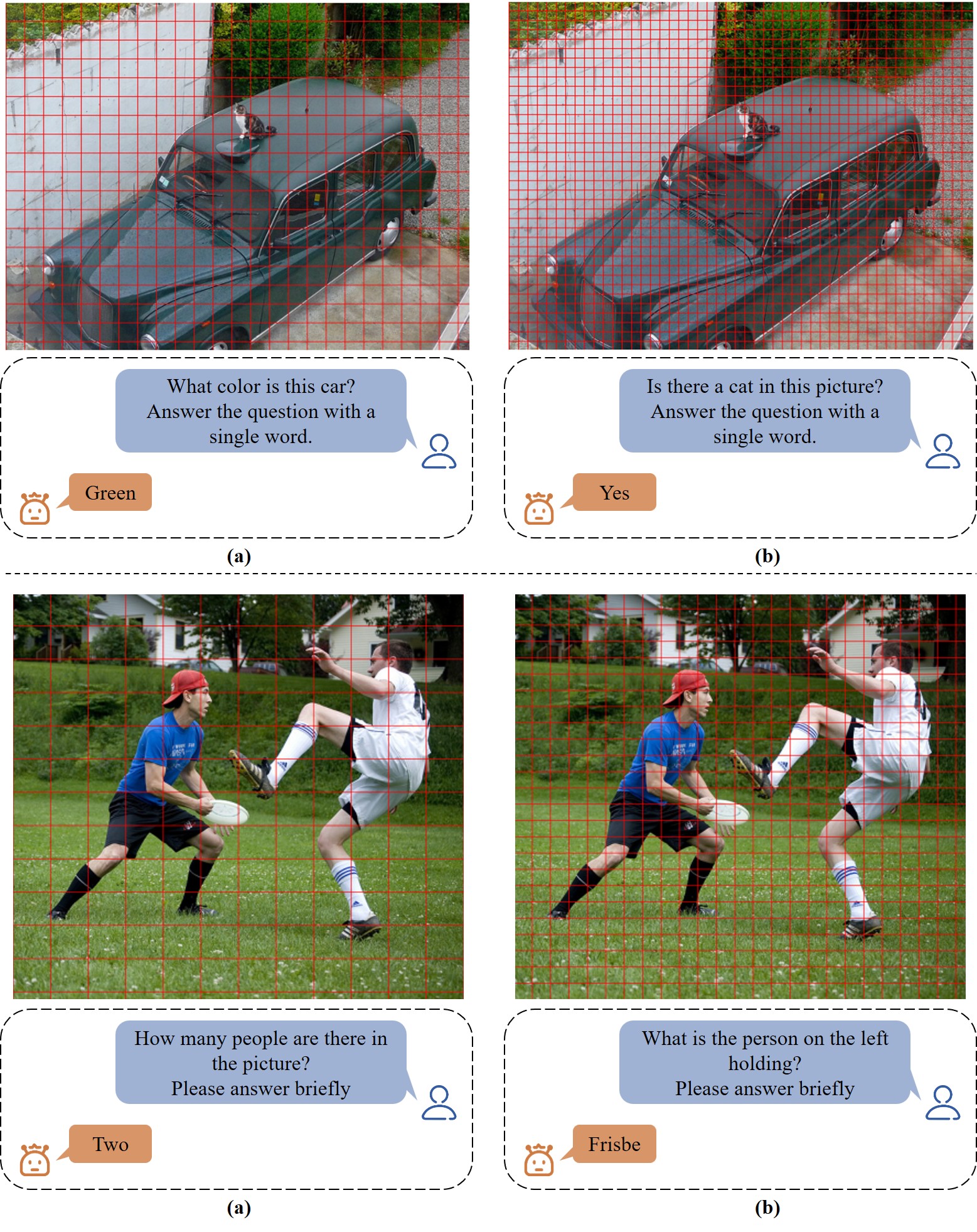}
    \caption{Visualization of granularity selection using different instructions.}
    \label{fig:grid}
\end{figure*}

As shown in Figure \ref{fig:grid}, we input the same image with different instructions and then visualize the selected visual granularity on the image, i.e., the number of patches. As can be seen, even for the same image, the router selects different visual granularities for different instructions. When asking about the color of the car, the model does not require such fine-grained visual information, whereas when asking whether there is a cat, the model requires finer-grained visual information.


\end{document}